\documentclass[letterpaper, 10 pt, conference]{ieeeconf}  

\IEEEoverridecommandlockouts                              

\usepackage{graphics} 
\usepackage{epsfig} 
\usepackage{amsmath} 
\usepackage{amssymb}  
\usepackage{algorithm,algorithmicx,algpseudocode} 
\algrenewcommand\algorithmicrequire{\textbf{Input:}}
\algrenewcommand\algorithmicensure{\textbf{Output:}}
\usepackage{bm} 

\usepackage[flushmargin,hang]{footmisc} 

\newcommand{\fig}[1]{Fig.~\ref{#1}}

\newcommand{\tab}[1]{Table~\ref{#1}}

\def\epsgaiji#1{\leavevmode\kern-0.025zw\raise-.37zh\hbox{%
  \epsfile{file=#1,width=1.05zw}}\kern-0.025zw}
\newcommand{\MARU}[1]{{\ooalign{\hfil#1\/\hfil\crcr\raise.167ex\hbox{\mathhexbox20D}}}}

\usepackage{array}

\newcommand{\Real}[1]{\mathbb{R}^{#1}}
\newcommand{\transpose}{^{\mathsf{T}}} 

\newcommand{\posVec}{\bm{p}}
\newcommand{\rotMat}{\bm{R}}

\newcommand{\graspablePoint}{\bm{g}}
\newcommand{\graspablePointsMap}{M}

\newcommand{\node}{v}
\newcommand{\nodeSet}{V}
\newcommand{\numNodes}{k}

\newcommand{\edge}{e}
\newcommand{\edgeSet}{E}
\newcommand{\numEdges}{l}

\newcommand{\graph}{G}

\newcommand{\idxBase}{\mathrm{b}}
\newcommand{\basePos}{\posVec_{\idxBase}}

\newcommand{\baseJacobi}{\bm{J}_{\mathrm{b}}}

\newcommand{\limbJacobiOf}[1]{\bm{J}_{\mathrm{m{#1}}}}

\newcommand{\baseTwist}{\bm{v}_{\mathrm{b}}}

\newcommand{\jointVelOf}[1]{\dot{\bm{q}}_{\mathrm{j}{#1}}}

\newcommand{\eeTwistOf}[1]{\bm{v}_{\mathrm{e{#1}}}}

\newcommand{\argmin}{\mathop{\rm arg~min}\limits}
\let\labelindent\relax
\usepackage{enumitem}

\usepackage{flushend} 
\usepackage{comment}

\usepackage{siunitx} 
\usepackage{url}

\usepackage{pgfplots}
\pgfplotsset{compat=newest}
\usetikzlibrary{plotmarks}
\usetikzlibrary{arrows.meta}
\usepgfplotslibrary{patchplots}
\usepackage{booktabs} 
\usepackage{multirow} 
\usepackage{grffile}
\pgfplotsset{plot coordinates/math parser=false}
\newlength\fwidth
\newlength\fheight

\usepackage{cite}

\title{\LARGE \bf
Graph-Based Simultaneous Path and Foothold Planning for Multi-Limbed Intra-Vehicular Robots in Space Stations
}

\author{Masazumi Imai$^{1}$, Kentaro Uno$^{2}$, Toshinori Kuwahara$^{3}$, and Kazuya Yoshida$^{2}$
\thanks{$^{*}$This work was supported by the Japan Society for the Promotion of Science (JSPS) KAKENHI under Grant JP25KJ0592.}%
\thanks{$^{1}$M. Imai is with the Department of Aerospace Engineering, Graduate School of Engineering, Tohoku University, Sendai 980-8579, Japan.}%
\thanks{$^{2}$K. Uno and K. Yoshida are with the New Industry Creation Hatchery Center (NICHe), Tohoku University, Sendai 980-8579, Japan.}%
\thanks{$^{3}$T. Kuwahara is with the Research Center for Green X-Tech and Research Center for Space Cross-Tech, Green Goals Initiative, Tohoku University, Sendai 980-8579, Japan.}%
\thanks{
\textit{The corresponding author is Masazumi Imai.} (\tt{imai.masazumi.p2@dc.tohoku.ac.jp})
    }%
}%

\begin{document}

\maketitle
\thispagestyle{empty}
\pagestyle{empty}


\begin{abstract}
Robot-aided operations in space stations are essential for reducing the workload of astronauts and improving the efficiency of on-orbit activities. Multi-limbed intra-vehicular robots (MLIVRs) equipped with grappling end-effectors have emerged as a promising solution, as they can securely grasp pre-existing interfaces, such as handrails and seat tracks, thereby enabling stable locomotion and forceful manipulation in microgravity environments. Since graspable locations on these interfaces are spatially limited and discretely distributed, motion planning for MLIVRs must be addressed jointly with foothold planning. This paper presents a simultaneous path and foothold planning framework based on graph theory for MLIVRs. The proposed method efficiently searches for feasible stance sequences for a multi-limbed robot while satisfying manipulability constraints. The effectiveness of the proposed framework is validated through simulations in a 3D model of the International Space Station (ISS) cabin, demonstrating its capability to generate feasible and efficient locomotion plans in realistic intra-vehicular environments.
\end{abstract}

\section{Introduction}\label{introduction}
While robotic technologies have been extensively employed in planetary exploration missions, their utilization in on-orbit human spaceflight operations remains limited. On the International Space Station (ISS), robotic systems have long been used to assist extra-vehicular activities (EVAs) through teleoperated manipulators, thereby reducing the risks associated with astronaut operations. In contrast, most intra-vehicular activities (IVAs) continue to rely heavily on crew labor, highlighting the need for more capable robotic assistants within the station. Such systems are anticipated to reduce crew workload and improve the utilization of limited crew time available aboard the station~\cite{ISS_Crew-time}.

To date, several relatively simple and repetitive tasks inside the ISS have been supported by free-flying robotic systems, including Astrobee by NASA~\cite{smith2026astrobee}, CIMON by DLR~\cite{eisenberg2025cimon}, and Int-Ball2 by JAXA~\cite{hirano2024intball}. Examples of their applications include recording experimental activities, providing additional third-person viewpoints during hands-on operations, and facilitating communication between the ISS crew and ground control centers~\cite{yamaguchi2025free}. However, other routine intra-vehicular tasks, including onboard logistics reconfiguration and experimental equipment setup, continue to be performed by crew members. These tasks often involve the transportation and deployment of large objects and require the generation of substantial interaction forces with the surrounding environment, making them difficult for free-flying robots to accomplish.
Therefore, more capable mobile robotic systems are required to perform such force-intensive tasks and reduce crew workload. The development of advanced robotic assistants is anticipated to play a key role in enabling sustainable and efficient operations aboard future crewed space stations and long-duration orbital habitats.
\begin{figure}[t]
\centerline{\includegraphics[width=\linewidth]{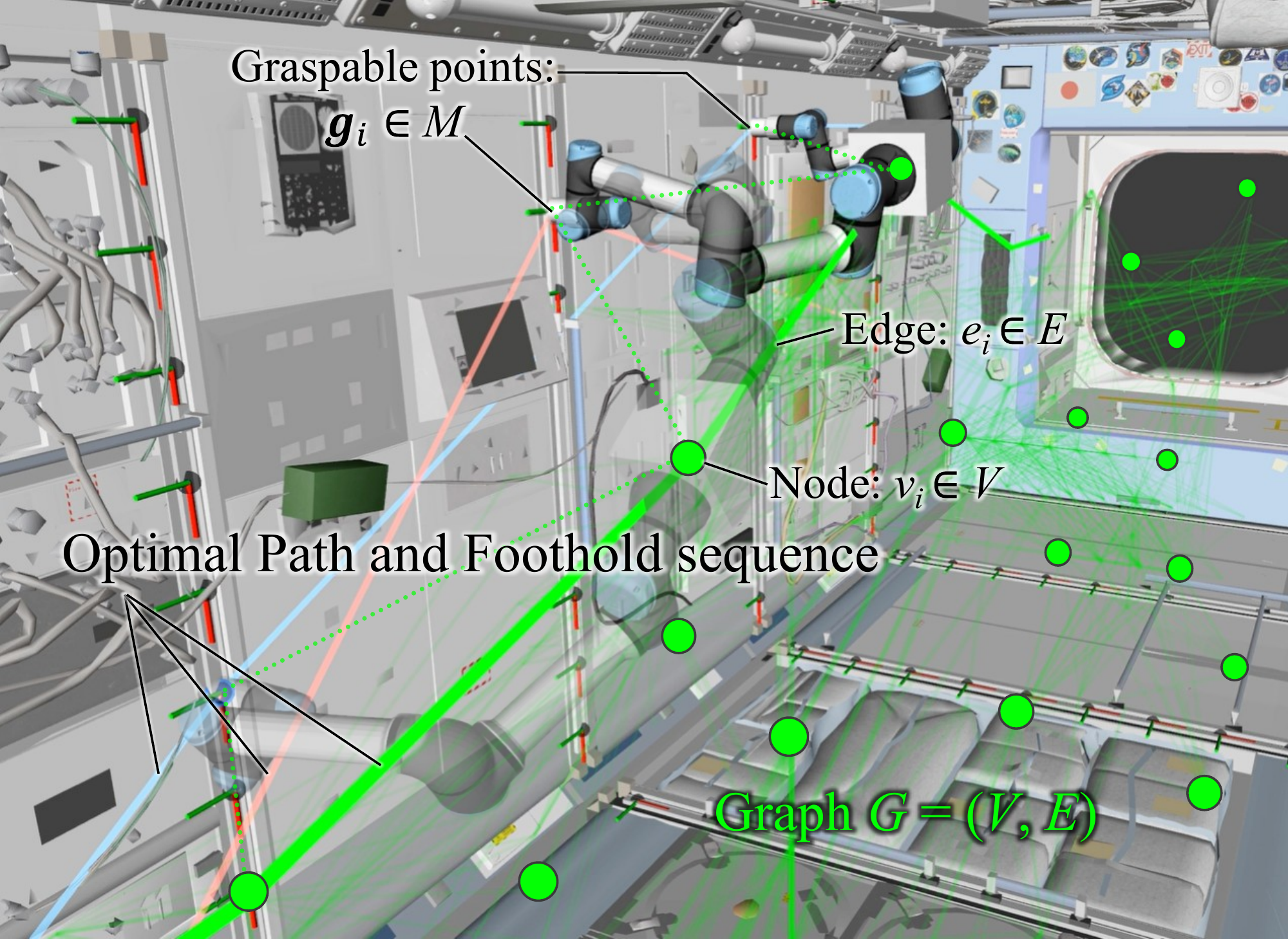}}
\caption{Rail-gripping locomotion of a multi-limbed intra-vehicular robot (MLIVR) in a space-station environment. Grasping locations (footholds) are discretely distributed along rails, such as seat tracks and handrails, inside the station module. The proposed planner simultaneously determines the optimal base trajectory (thick green line) and foothold sequence (red and blue lines) from all feasible locomotion paths (thin green lines) represented in a graph structure. Candidate stances that violate kinematic constraints or exhibit insufficient manipulability are excluded from the graph, thereby improving computational efficiency and motion robustness.
}
\label{fig1}
\end{figure}

Mobile manipulators equipped with grappling end-effectors at both ends have emerged as a promising solution for performing capture, manipulation, and docking tasks in microgravity environments. This concept has long been employed for external space station operations, as exemplified by the Space Station Remote Manipulator System (SSRMS), also known as Canadarm2~\cite{ssrms}, and the European Robotic Arm (ERA)~\cite{era} on the ISS. Similar systems have also been deployed on the Chinese Space Station, including the Core Space Station Cabin Manipulator (CSSCM) and the Experimental Space Station Cabin Manipulator (ESSCM)~\cite{cssm}. 
All of these systems feature an ``inchworm-like'' locomotion capability, whereby they move by sequentially attaching to dedicated grapple fixtures. This capability enables the manipulator to relocate its base across large structures while maintaining a stable mechanical connection to the environment, thereby significantly extending its operational workspace.

Likewise, this environment-gripping locomotion capability can also be employed for an intra-vehicular robot to stably perform the force-demanding tasks in microgravity. We refer to such a robot as a {\it Multi-Limbed Intra-Vehicular Robot (MLIVR)}. Regarding the grasping targets in the interior of the space station, instead of newly installing the dedicated grapple fixtures in the cabin, it is more efficient to grasp pre-existing standardized interfaces, such as astronaut handrails and seat tracks, is practically efficient. So far, several demonstrations of such ISS rail-gripping locomotion have been conducted by a humanoid robot~\cite{R2_mobility} and a multi-limbed robot~\cite{yamaguchi2025towards}. 

For a MLIVR, graspable locations are spatially limited and discretely distributed along the rails installed inside space station modules. 
As a result, locomotion planning for MLIVRs is inherently coupled with foothold selection, requiring the robot to determine both a feasible route and an appropriate sequence of grasping locations simultaneously.

\subsection{Related Work}
Research on foothold planning can be found in the field of legged climbing robots.
Starting from the pioneering work of Bretl {\it et al.}~\cite{bretl2006motion}, numerous studies on wall-climbing and cliff-climbing robots have investigated the problem of selecting suitable gripping locations while satisfying constraints such as static stability and kinematic reachability~\cite{albee2019motion,uno2019gait}. 
However, these studies generally assume that graspable locations are distributed over a two-dimensional surface and determine footholds in a local and incremental manner. As a result, the overall route emerges from a series of local decisions rather than being optimized globally. Such an assumption is unsuitable for intra-vehicular locomotion, where graspable interfaces are sparsely and discretely distributed in three-dimensional space. Consequently, foothold selection and global path planning must be considered simultaneously to generate efficient and feasible locomotion plans.

In 2022, Xu {\it et al.}~\cite{xu2021contact} and, subsequently, Takada {\it et al.}~\cite{takada2023graph}  in 2023 proposed search-based planning frameworks for multi-limbed robots operating under gravity.
The former simultaneously plans gait and foothold sequences using Monte Carlo tree search, whereas the latter simultaneously plans locomotion paths and foothold sequences using graph search.
Takada {\it et al.}'s method represents kinematically reachable combinations of footholds as graph nodes and determines feasible locomotion plans through graph search.
In 2024, Rodriguez {\it et al.}~\cite{rodriguez2024hybrid} presented a hybrid planning algorithm that generates kinematically feasible step sequences for loco-manipulation tasks performed by a multi-limbed orbital robot. While demonstrating a practical implementation framework, the study was also limited to two-dimensional problem settings and provided relatively limited discussion on the mathematical formulation of the foothold planning problem.


\subsection{Contribution}
While several related studies have been reported, a significant gap remains between the previously proposed discrete foothold sequence planning frameworks and their practical application to three-dimensional intra-vehicular locomotion in space stations. To bridge this gap, this paper presents a graph-based framework for the simultaneous planning of locomotion paths and footholds in three-dimensional intra-vehicular environments (see \fig{fig1}). Building upon the method proposed in~\cite{takada2023graph}, the framework is extended to account for spatial locomotion and practical constraints related to robot motion feasibility.

The proposed method explicitly considers the discrete distribution of graspable interfaces in space stations and generates locomotion plans for multi-limbed intravehicular robots while satisfying kinematic feasibility constraints and maintaining manipulability above a predefined threshold. Furthermore, several cost functions for the A$^*$-based graph search are investigated and compared to identify computationally efficient and practically effective planning settings. 

Finally, the proposed framework is validated through a dynamic simulation in a realistic ISS-module environment, allowing for evaluation of physical performance metric for locomotion trajectories generated by the planner under microgravity conditions. The results highlight the potential of graph-based planning as a practical solution for robotic systems capable of autonomously performing routine intra-vehicular tasks in future orbital habitats.


\section{Method}

For a multi-limbed robot operating in an environment where graspable points are sparse and discretely distributed, the simultaneous planning of locomotion paths and footholds is formulated as a graph-search problem, as illustrated in \fig{fig1}.
The pseudocode for the proposed planning method is also summarized in Algorithm~\ref{alg:path_and_foothold_planning}. It is important to note that this method is generally applicable to an $N$-limbed robot.

\subsection{Problem Formulation by Graph Representation}

Let the environment be represented as a map $\graspablePointsMap$ containing $m$ discrete graspable points,
\begin{equation}
    \graspablePointsMap
    =
    \left\{
        \graspablePoint_{1},
        \graspablePoint_{2},
        \dots,
        \graspablePoint_{m}
    \right\}.
\end{equation}
The $i$-th graspable point is defined as $\graspablePoint_i=(\posVec_i,\rotMat_i)$, where $\posVec_i\in\Real{3}$ denotes its position and $\rotMat_i\in SO(3)$ denotes the orientation of the local grasp frame.

A stance of the robot, corresponding to a node $\node_{i}$, is defined as an ordered tuple of the graspable points assigned to the robot's $N$ limbs:
\begin{equation} \label{eq:node_vj}
    \node_{i}
    =
    \left(
        \graspablePoint_{i,1}, \,
        \graspablePoint_{i,2}, \,
        \dots , \,
        \graspablePoint_{i,N}
    \right),
    \quad
    \graspablePoint_{i,j} \in \graspablePointsMap,
\end{equation}
where $\graspablePoint_{i,j}$ denotes the graspable point assigned to the $j$-th limb in stance $\node_i$.

Let $\nodeSet$ denote the set of all feasible stances. If $\numNodes$ feasible stances exist in the map, $\nodeSet$ is given by
\begin{equation} \label{eq:node_set}
    \nodeSet
    =
    \left\{
        \node_{1}, \,
        \node_{2}, \,
        \dots , \,
        \node_{\numNodes}
    \right\}.
\end{equation}

An edge represents a single feasible locomotion step from one stance to another.
If the transition from a node $\node_{i}$ to $\node_{j}$ is feasible, an edge $\edge = (\node_{i}, \, \node_{j})$ is established between them.
If $\numEdges$ feasible transitions exist, the edge set $\edgeSet$ is given by
\begin{equation}
    \edgeSet
    =
    \left\{
        \edge_{1}, \,
        \edge_{2}, \,
        \dots , \,
        \edge_{\numEdges}
    \right\}.
    \label{eq:edge_set}
\end{equation}

The resulting graph representing the feasible robot stances and transitions on the discrete graspable points is then defined as $\graph=(\nodeSet,\edgeSet)$.

A graph-search algorithm, such as Dijkstra's algorithm or A$^*$, can then be used to obtain a globally cost-optimal path from the start stance toward the goal.
Let $\node_{\mathrm{s}}$ denote the start node and
$\node_{\mathrm{g}}$ denote the terminal node satisfying the goal condition.
A path containing $L$ locomotion steps is represented by the ordered node sequence
\begin{equation}
    P
    =
    \left(
        \node_{r_0}, \,
        \node_{r_1}, \,
        \dots, \,
        \node_{r_L}
    \right),
    \quad
    \node_{r_0}=\node_{\mathrm{s}},
    \
    \node_{r_L}=\node_{\mathrm{g}},
\end{equation}
where $\node_{r_i}\in\nodeSet$ and $(\node_{r_{i-1}},\node_{r_i})\in\edgeSet$ for $i=1,\dots,L$.
The goal is considered reached when $\|\basePos(\node)-\posVec_{\mathrm{g}}\| \leq \epsilon_{\mathrm{goal}}$, where $\basePos(\node)$ denotes the base position associated with stance $\node$ and $\epsilon_{\mathrm{goal}}$ is the goal tolerance.

From \eqref{eq:node_vj}, the corresponding foothold sequence of the $i$-th limb is
\begin{equation}
    F_{\mathrm{s \to g}, i}
    =
    \left(
        \graspablePoint_{r_0,i}, \,
        \graspablePoint_{r_1,i}, \,
        \dots, \,
        \graspablePoint_{r_L,i}
    \right).
\end{equation}
The foothold sequences for all $N$ limbs are collectively denoted by
\begin{equation}
    F_{\mathrm{s}\to\mathrm{g}}
    =
    \left(
        F_{\mathrm{s}\to\mathrm{g},1},
        \dots,
        F_{\mathrm{s}\to\mathrm{g},N}
    \right).
\end{equation}

\subsection{Kinematic Feasibility and Manipulability Incorporation} \label{seq:kin_mani}

When considering path and foothold planning, whether the robot satisfies kinematic feasibility is a crucial metric. 
Accordingly, in our previous work~\cite{takada2023graph}, a two-step kinematic validation was incorporated into the graph construction process. First, geometric reachability based on the maximum workspace limits of the limbs was evaluated to efficiently eliminate unreachable nodes. Second, inverse kinematics (IK) was solved to select only feasible stances.
However, in microgravity or steep climbing environments, simple geometric reachability is insufficient to guarantee safe locomotion.
The robot must maintain a sufficient margin against kinematic singularities to manage reaction forces.
Therefore, we introduce an additional constraint using not only IK but also two distinct manipulability measures: limb manipulability and base manipulability for more robust motion.

Since a spatial Jacobian contains translational and rotational components with different physical dimensions, directly applying a manipulability measure to the unscaled Jacobian results in values that depend on the relative scaling of these components.
To obtain dimensionally consistent manipulability measures, we employ characteristic-length normalization~\cite{stocco1999use}.
Let $\ell_{\mathrm{c}}$ denote the characteristic length, defined in this study as the maximum kinematic reach of a limb.
For a spatial velocity ordered as linear and angular components, the scaling matrix is defined as
\begin{equation}
  \bm{S}_{\mathrm{c}}
  =
  \begin{bmatrix}
    \ell_{\mathrm{c}}^{-1}\bm{I}_{3} & \bm{0} \\
    \bm{0} & \bm{I}_{3}
  \end{bmatrix}.
\end{equation}
where $\bm{I}_{d}$ denotes the $d \times d$ identity matrix.
Accordingly, for any spatial Jacobian $\bm{J}\in\mathbb{R}^{6\times n}$, its normalized form is defined as
\begin{equation}
  \bar{\bm{J}}
  =
  \bm{S}_{\mathrm{c}}\bm{J}.
\end{equation}
The same characteristic length is used for both the limb and base manipulability measures throughout the planning process.
This normalization removes the dependence on the choice of length units and, for the all-revolute limbs considered in this study, yields dimensionless manipulability measures.

\subsubsection{Limb Manipulability}

For the swing limb reaching the next graspable point, we evaluate the velocity manipulability measure $w_{\mathrm{m}_{i}}$~\cite{yoshikawa1985manipulability} of the $i$-th limb.
Let $\limbJacobiOf{_{\it{i}}} \in \mathbb{R}^{6 \times n}$ denote the spatial Jacobian of the limb with $n$ joints.
Using the characteristic-length scaling defined above, the normalized limb Jacobian is given by
\begin{equation}
  \bar{\bm{J}}_{\mathrm{m}_{i}}
  =
  \bm{S}_{\mathrm{c}} \limbJacobiOf{_{\it{i}}}.
\end{equation}
The limb manipulability measure is then defined as
\begin{equation}
  w_{\mathrm{m}_{i}}
  =
  \sqrt{
    \det
    \left(
      \bar{\bm{J}}_{\mathrm{m}_{i}}
      \bar{\bm{J}}_{\mathrm{m}_{i}}^{\hspace{0.1em}\transpose}
    \right)
  }.
  \label{eq:w_limb}
\end{equation}
A low $w_{\mathrm{m}_{i}}$ indicates that the limb is close to a kinematic singularity, reducing its capability to generate end-effector motion while approaching the target.
Therefore, let $\epsilon_{\mathrm{m}}$ be the threshold for the normalized limb manipulability measure; during graph generation, any stance with $w_{\mathrm{m}_{i}} < \epsilon_{\mathrm{m}}$ is considered unsafe and removed from the graph.

\subsubsection{Base Manipulability}
For multi-limbed climbing robots equipped with grippers at each limb tip, the mobility of their floating base is constrained by the closed-chain kinematics formed by grasping the environment with the supporting limbs.
For such robots, evaluating only single-limb manipulability cannot accurately assess the mobility of the whole system.
The situation where multiple limbs form a closed kinematic chain with the environment can be considered equivalent to the cooperative manipulation of an object by multiple robotic arms.
For the cooperative control of multiple arms, methods to evaluate the manipulability of multi-arm systems have been proposed in~\cite{chiacchio1991global} by extending the definition of the single-arm manipulability measure in \eqref{eq:w_limb}.
Therefore, we apply this concept to multi-limbed climbing robots to introduce a method for evaluating the base mobility of this type of robot.

Let $\eeTwistOf{_\mathrm{sup}} \in \mathbb{R}^{6k}$ denote the stacked spatial velocities of the $k$ contact points associated with the supporting limbs.
Under the assumed rigid 6-DoF grasp, the supporting end-effectors are fixed to the environment, and hence $\eeTwistOf{_\mathrm{sup}} = \bm{0}$.
The resulting closed-chain velocity constraint is given by
\begin{equation}
  \baseJacobi \baseTwist
  +
  \limbJacobiOf{_\mathrm{sup}} \jointVelOf{_\mathrm{sup}}
  =
  \bm{0},
  \label{eq:closed_chain}
\end{equation}
where $\baseJacobi \in \mathbb{R}^{6k\times6}$ is the stacked base Jacobian and $\limbJacobiOf{_\mathrm{sup}} \in \mathbb{R}^{6k\times n_{\mathrm{sup}}}$ is the stacked joint Jacobian of the supporting limbs.
Under a rigid 6-DoF grasp, each $6\times6$ block of $\baseJacobi$ associated with a supporting contact represents a nonsingular spatial-velocity transformation.
Therefore, $\baseJacobi$ has full column rank, i.e., $\operatorname{rank}(\baseJacobi)=6$.

For multiple simultaneous contacts, however, an arbitrary supporting-limb joint velocity does not necessarily satisfy the closed-chain constraint.
An exact solution of \eqref{eq:closed_chain} exists only if
\begin{equation}
  \left(
    \bm{I}_{6k}
    -
    \baseJacobi\baseJacobi^{\dagger}
  \right)
  \limbJacobiOf{_\mathrm{sup}} \jointVelOf{_\mathrm{sup}}
  =
  \bm{0},
  \label{eq:consistency}
\end{equation}
Let $\bm{N}_{\mathrm{c}} \in \mathbb{R}^{n_{\mathrm{sup}}\times r_\mathrm{c}}$ be an orthonormal basis of
$
  \mathcal{N}
  \left[
    \left(
      \bm{I}_{6k}
      -
      \baseJacobi\baseJacobi^{\dagger}
    \right)
    \limbJacobiOf{_\mathrm{sup}}
  \right],
$
where $\mathcal{N}(\cdot)$ denotes the null space and $r_\mathrm{c}$ is the dimension of the constraint-consistent joint-velocity subspace.
The admissible joint velocities can then be parameterized as
$
  \jointVelOf{_\mathrm{sup}} = \bm{N}_{\mathrm{c}} \bm{\nu}_{\mathrm{c}},
$
where $\bm{\nu}_{\mathrm{c}} \in \mathbb{R}^{r_\mathrm{c}}$ denotes the independent velocity coordinates spanning the constraint-consistent subspace.
Since $\baseJacobi$ has full column rank, substitution into \eqref{eq:closed_chain} yields the unique base velocity
\begin{align}
  \baseTwist
  &=
  -\baseJacobi^{\dagger}
  \limbJacobiOf{_\mathrm{sup}}
  \bm{N}_{\mathrm{c}} \bm{\nu}_{\mathrm{c}} \notag\\
  &=
  \bm{J}_{\mathrm{eq}} \bm{\nu}_{\mathrm{c}},
  \qquad\qquad
  \bm{J}_{\mathrm{eq}}
  =
  -\baseJacobi^{\dagger}
  \limbJacobiOf{_\mathrm{sup}}
  \bm{N}_{\mathrm{c}},
  \label{eq:equivalent_jacobian}
\end{align}
where $\bm{J}_{\mathrm{eq}} \in \mathbb{R}^{6\times r_\mathrm{c}}$ is the constraint-consistent equivalent Jacobian.
Because $\bm{N}_{\mathrm{c}}$ is orthonormal, $\|\jointVelOf{_\mathrm{sup}}\| = \|\bm{\nu}_{\mathrm{c}}\|$.
To account for the different physical dimensions of the translational and rotational components of the base spatial velocity, the equivalent Jacobian is normalized using the same characteristic-length scaling as that used for the limb manipulability:
\begin{equation}
  \bar{\bm{J}}_{\mathrm{eq}}
  =
  \bm{S}_{\mathrm{c}}
  \bm{J}_{\mathrm{eq}}.
\end{equation}
Accordingly, analogous to the conventional velocity manipulability measure, the normalized base manipulability is defined as
\begin{equation}
  w_{\mathrm{b}}
  =
  \sqrt{
    \det
    \left(
      \bar{\bm{J}}_{\mathrm{eq}}
      \bar{\bm{J}}_{\mathrm{eq}}^{\hspace{0.1em}\transpose}
    \right)
  }.
  \label{eq:base_manipulability}
\end{equation}
This measure quantifies the robot's capability to generate base spatial motion while satisfying the closed-chain constraints.
Let $\epsilon_{\mathrm{b}}$ be the threshold for the normalized base manipulability measure.
During graph construction, as with the limb manipulability measure, stances satisfying $w_{\mathrm{b}} < \epsilon_{\mathrm{b}}$ are not selected as nodes.

\begin{algorithm}[t]
    \caption{Graph-based Path and Foothold Planning}
    \label{alg:path_and_foothold_planning}
    \begin{algorithmic}[1]
        \Require
            \Statex $\graspablePointsMap$: Map of graspable points
            \Statex $\node_{\mathrm{s}}$: Start stance
            \Statex $\posVec_{\mathrm{g}}$: Goal position
        \Ensure
            \Statex $P$: Globally cost-optimal stance path
            \Statex $F_{\mathrm{s \to g}}$: Foothold sequences
        
        \Function{PathAndFootholdPlanner}{$\graspablePointsMap$, $\node_{\mathrm{s}}$, $\posVec_{\mathrm{g}}$}

            \State $\mathcal{O} \gets \{\node_{\mathrm{s}}\}$,
                   $\mathcal{C} \gets \emptyset$
            \State $g(\node_{\mathrm{s}}) \gets 0$
            \State $f(\node_{\mathrm{s}}) \gets h(\node_{\mathrm{s}},\posVec_{\mathrm{g}})$
            
            \While{$\mathcal{O} \neq \emptyset$}

                \State $\node_{\mathrm{curr}} \gets \argmin_{\node \in \mathcal{O}} f(\node)$
                
                \If{$\| \posVec_{\idxBase}(\node_{\mathrm{curr}}) - \posVec_{\mathrm{g}} \| \leq \epsilon_{\mathrm{goal}}$}
                    \State $P \gets \Call{ConstructPath}{\node_{\mathrm{curr}}}$
                    \State $F_{\mathrm{s}\to\mathrm{g}} \gets \Call{ExtractFootholds}{P}$

                    \State \Return $(P,F_{\mathrm{s}\to\mathrm{g}})$
                \EndIf
                
                \State $\mathcal{O} \gets \mathcal{O} \setminus \{ \node_{\mathrm{curr}} \}$, \ 
                       $\mathcal{C} \gets \mathcal{C} \cup \{ \node_{\mathrm{curr}} \}$
                
                \ForAll{$\node_{\mathrm{next}} \in \operatorname{Nbr}(\node_{\mathrm{curr}}) \setminus \mathcal{C}$}

                    \If{\Call{IsKinematicsFeasible}{$\node_{\mathrm{next}}$}
                        \textbf{and} \\ \hspace{5.1em}
                        \Call{IsManipulabilityOK}{$\node_{\mathrm{next}}$}
                    }

                        \State $g_{\mathrm{new}} \gets g(\node_{\mathrm{curr}}) + c(\node_{\mathrm{curr}}, \node_{\mathrm{next}})$
                        
                        \If{
                            $\node_{\mathrm{next}} \notin \mathcal{O}$
                            \textbf{or}
                            $g_{\mathrm{new}} < g(\node_{\mathrm{next}})$
                        }

                            \State $g(\node_{\mathrm{next}}) \gets g_{\mathrm{new}}$
                            \State $f(\node_{\mathrm{next}}) \gets g_{\mathrm{new}} + h(\node_{\mathrm{next}}, \posVec_{\mathrm{g}})$
                            \State $\mathrm{parent}(\node_{\mathrm{next}}) \gets \node_{\mathrm{curr}}$
                            \State $\mathcal{O} \gets \mathcal{O} \cup \{ \node_{\mathrm{next}} \}$

                        \EndIf
                    \EndIf
                \EndFor
            \EndWhile
            
            \State \Return Failure

        \EndFunction
    \end{algorithmic}
\end{algorithm}
\subsection{Cost Functions for Locomotion Evaluation} \label{seq:cost_func}

The planning process is performed using the A$^*$ algorithm, which searches for a cost-optimal path according to the defined evaluation criteria. The cost function $f$ at the $i$-th node $v_i$ is generally defined as the sum of the accumulated actual cost $g$ and the heuristic cost $h$:
\begin{equation}
    f(v_i) = g(v_i) + h(v_i).
\end{equation}

To evaluate the impact of different locomotion strategies, the transition cost between nodes is defined selectively. 
In this study, we introduce two independent cost functions. By minimizing each cost individually during the A$^*$ search, we can quantitatively compare the characteristics of the resulting paths.

\subsubsection{Base Translation Cost}

This cost focuses on minimizing the travel distance of the robot's base.
\begin{align} \label{eq:base_trans_cost}
    f_{\mathrm{base}}(\node_{i}) = \; & g_{\mathrm{base}}(\node_{i}) + h_{\mathrm{base}}(\node_{i}), \notag
    \\
    g_{\mathrm{base}}(\node_{i}) \; &= \sum_{j = 1}^{i} \| \basePos(\node_{j}) - \basePos(\node_{j-1}) \|, \notag
    \\
    h_{\mathrm{base}}(\node_{i}) \; &= \| \basePos(\node_{i}) - \posVec_{\mathrm{g}} \|.
\end{align}

\subsubsection{Step Count Cost}

This cost assigns a constant penalty for each transition, thereby encouraging the robot to maximize its reach and minimize the total number of grasping maneuvers.
\begin{align} \label{eq:step_count_cost}
    f_{\mathrm{step}}(\node_{i}) = \; & g_{\mathrm{step}}(\node_{i}) + h_{\mathrm{step}}(\node_{i}), \notag
    \\
    g_{\mathrm{step}}(\node_{i}) \; & =  \sum_{j=1}^{i} n_{\mathrm{step},j}, \notag
    \\
    h_{\mathrm{step}}(\node_{i}) \; & = \frac{\| \basePos(\node_{i}) - \posVec_{\mathrm{g}} \|}{d_{\max}}.
\end{align}
where $n_{\mathrm{step},i}$ is the number of steps at the $i$-th node, and $d_{\max}$ is the maximum movement of the base in one step.

Here, $\mathcal{O}$ and $\mathcal{C}$ denote the open and closed sets used in the graph search, respectively.
For a node $\node_i$, $\operatorname{Nbr}(\node_i)$ denotes the set of candidate neighboring stances generated by relocating one limb to another graspable point.
The overall planning procedure, integrating the graph representation, kinematic-feasibility and manipulability constraints, and the A$^{*}$-based cost evaluation described above, is summarized in Algorithm~\ref{alg:path_and_foothold_planning}.

\section{Algorithm Evaluation} \label{sec:evaluation}
\begin{figure*}[t]
    \centering
    \footnotesize
    \begin{minipage}[t]{0.32\linewidth}
        \centering
        \includegraphics[width=\linewidth]{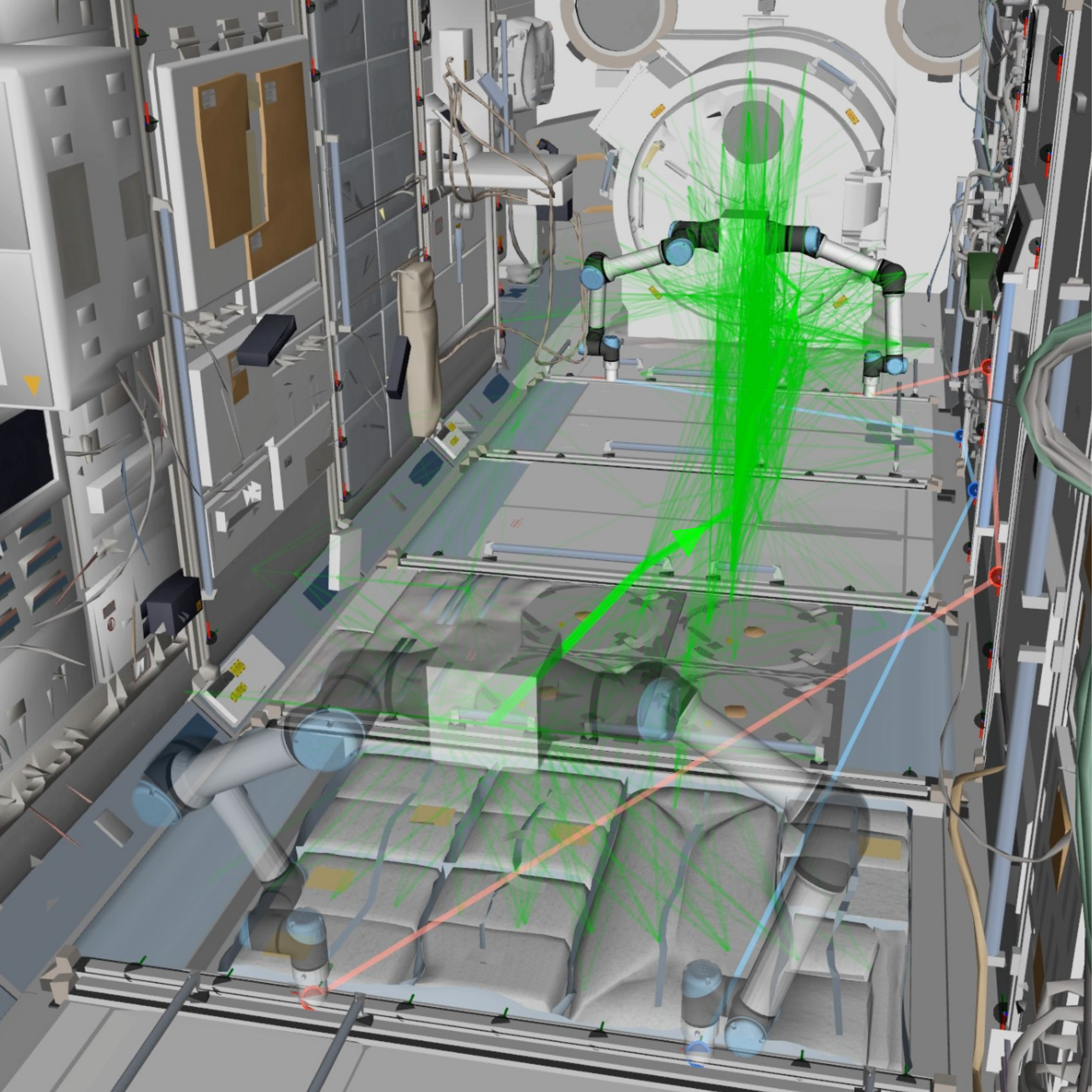}
        - Manipulability-based node selection: no\\
        - Cost: base translation\\
        \vspace{2mm}
        {\it Case 1} 
    \end{minipage}
    \hspace{0.005\linewidth}
    \begin{minipage}[t]{0.32\linewidth}
        \centering
        \includegraphics[width=\linewidth]{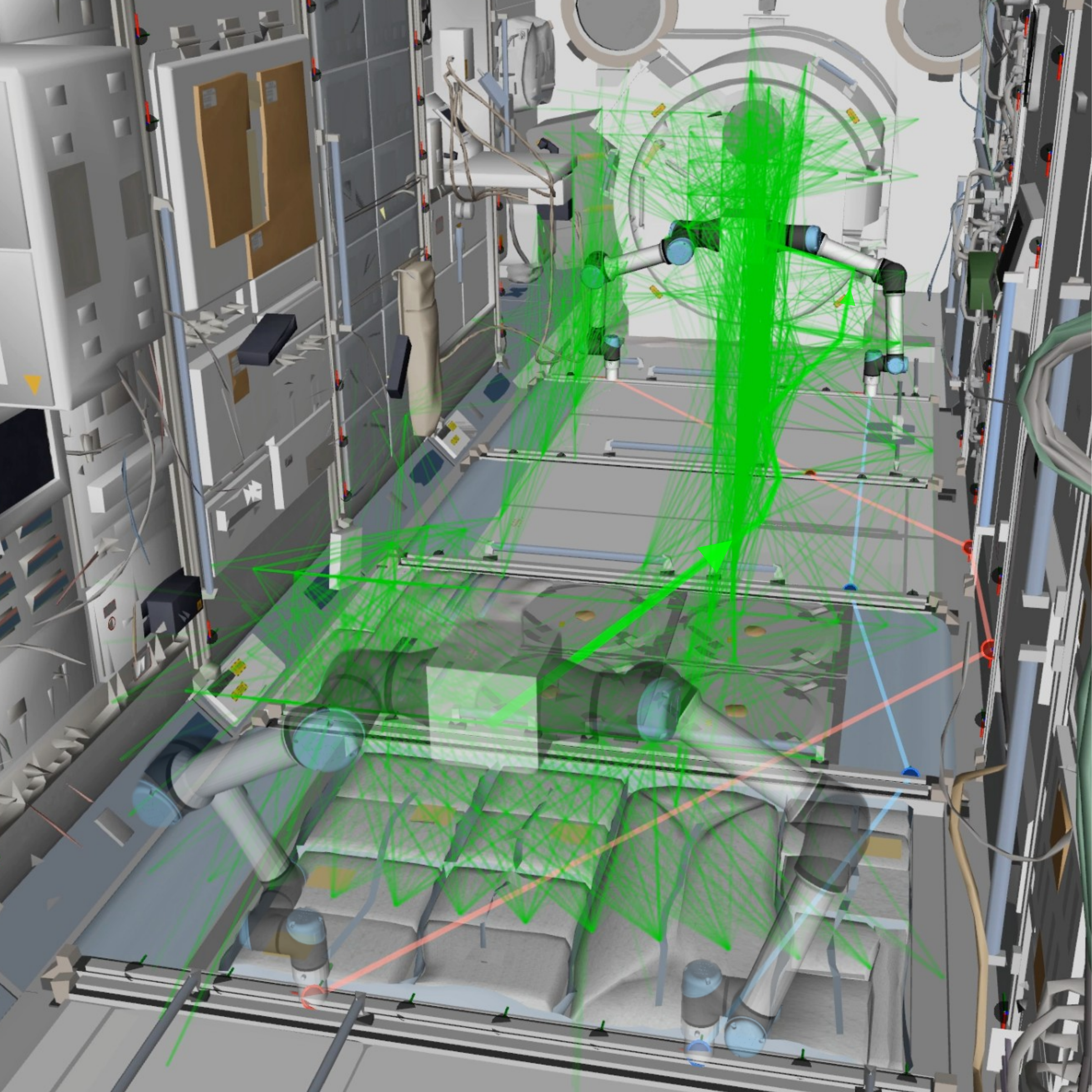}
        - Manipulability-based node selection: yes\\
        - Cost: base translation\\
        \vspace{2mm}
        {\it Case 2}
    \end{minipage}
    \hspace{0.005\linewidth}
    \begin{minipage}[t]{0.32\linewidth}
        \centering
        \includegraphics[width=\linewidth]{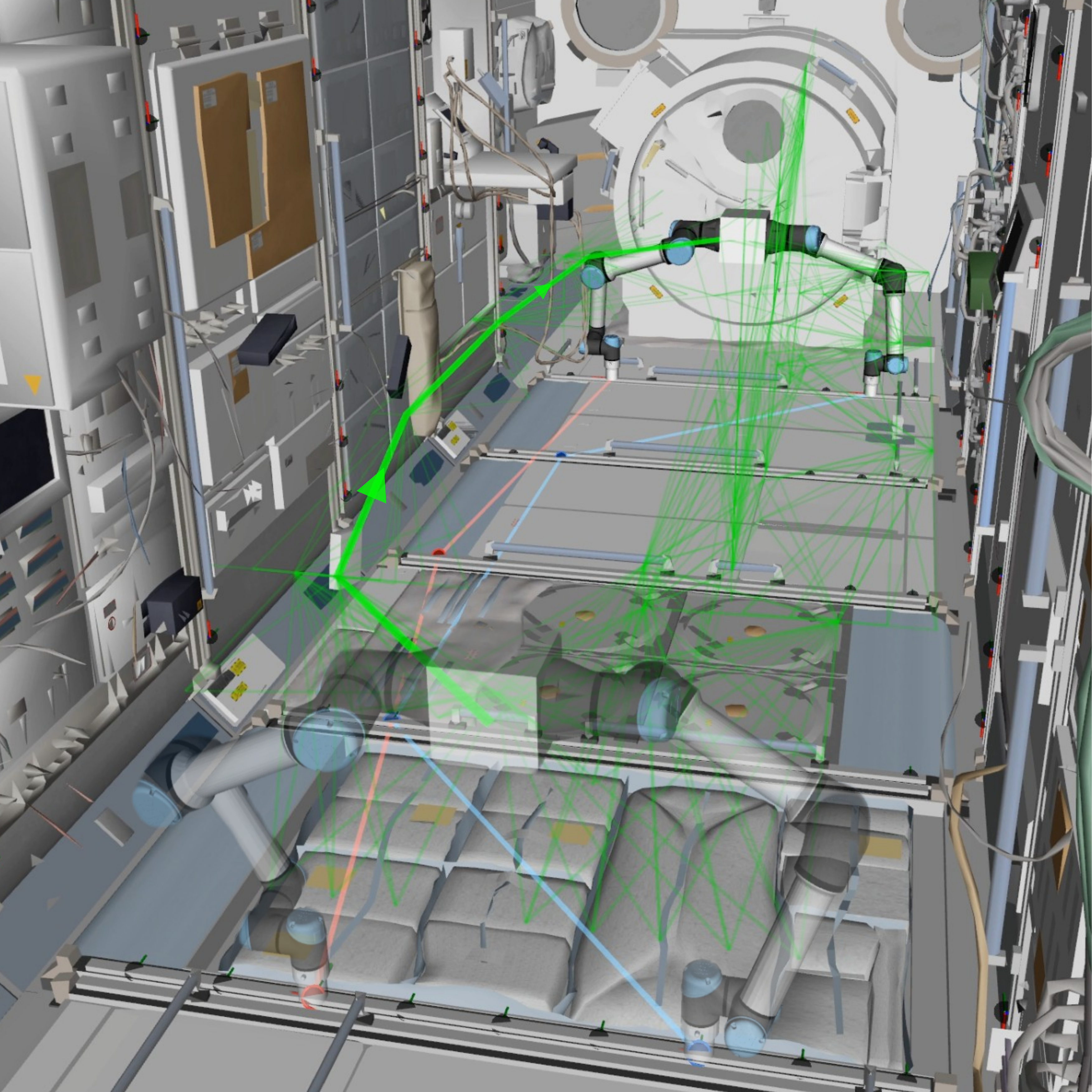}
        - Manipulability-based node selection: yes\\
        - Cost: step count\\
        \vspace{2mm}
        {\it Case 3}
    \end{minipage}
    \vspace{2mm}
    \caption{Cost-optimal paths and foothold sequences for a multi-limbed intra-vehicular robot on discrete graspable points in the ISS under different computation settings with A$^*$. The thin and thick green lines indicate the explored candidate base paths and the selected base path, respectively. The half-transparent red and blue lines represent the foothold sequences for the corresponding limbs.}
    \label{fig:case_study}
\end{figure*}
\begin{table*}[t]
    \centering
    \caption{Quantitative comparison among the simulation result cases.}
    \vspace{-1mm}
    \resizebox{\linewidth}{!}{
    \begin{tabular}{c | c  c | r r r r r r}
        \toprule
        \multirow{3}{*}{Case}
        & \multicolumn{2}{c|}{Algorithm Settings}
        & \multicolumn{6}{c}{Metrics for the Result Evaluation}
        \\
        & \multirow{1}{*}{Manipulability-based}
        & \multirow{2}{*}{Cost type}
        & \multirow{1}{*}{Explored}
        & \multicolumn{1}{c}{Total base}
        & \multirow{1}{*}{Total step}
        & \multirow{2}{*}{$w_{\mathrm{m},\min}$ [-]}
        & \multirow{2}{*}{$w_{\mathrm{b},\min}$ [-]}
        & \multicolumn{1}{c}{Computation}
        \\
        & \multirow{1}{*}{node selection}
        &
        & \multicolumn{1}{c}{nodes [-]}
        & \multicolumn{1}{c}{translation [m]}
        & count [-]
        &
        &
        & \multicolumn{1}{c}{time [s]}
        \\
        \midrule
        1 & no  & base translation & 172 & 4.4 & 6 & 0.07 & 0.0008 & 1.2 \\
        2 & yes & base translation & 386 & 4.5 & 7 & 0.10 & 0.0015 & 1.8 \\
        3 & yes & step count       & 40  & 4.7 & 5 & 0.12 & 0.0018 & 0.2 \\
        4 & yes & base translation & 72  & 5.0 & 7 & 0.12 & 0.0020 & 0.2 \\
        \bottomrule
    \end{tabular}
    }
    \label{tab:case_metrics}
\end{table*}
To evaluate the effectiveness of the proposed method, a comparative case study was conducted using different algorithm configurations. The study was performed in a kinematic simulation environment representing an intra-vehicular scenario.
The computing platform was a laptop equipped with an Intel Core i9-12900H processor and 64\;GB of memory.

\subsection{Problem Settings}

The simulated environment models the interior of the Japanese Experiment Module (JEM) on the ISS.
The seat tracks installed inside the module are targeted as graspable points. For planning purposes, continuous seat tracks were discretized at 0.2-m intervals to generate discrete graspable points. However, locations where handrails are installed were excluded from the list of graspable points. The positions of all graspable points are assumed to be known in advance.
Furthermore, we do not take into account collisions between the robot and its environment.

As a representative intra-vehicular mobile robot, a multi-limbed robot with limbs arranged in an axisymmetric configuration is considered, following concepts previously proposed for orbital applications~\cite{ohkami1999operational,hayashi2000design,deremetz2021concept,hoyt2013spiderfab,zhao2024model}.
Although robots with three and more limbs are generally envisioned for executing complex tasks, two limbs are sufficient to provide stable mobility in microgravity environments. Therefore, this study employs a 12-DoF floating-base robot model equipped with two symmetric limbs, where each limb consists of a 6-DoF robotic manipulator (UR5e). Each end-effector is equipped with a gripper to grapple the rails.
It is assumed that the robot can acquire its base pose via an IMU, and its joint states via encoders.

\subsection{Algorithm Settings}

To evaluate the effectiveness of the proposed path planning method and analyze the characteristics of the cost functions, we conducted comparative studies under the following conditions:
\begin{enumerate}[label={\it Case \arabic*:}, leftmargin=1.5cm]
  \item A baseline using the base translation cost in \eqref{eq:base_trans_cost} without manipulability-based node selection.
  \item Using the base translation cost with manipulability-based node selection.
  \item Using the step count cost in \eqref{eq:step_count_cost} with manipulability-based node selection.
\end{enumerate}
For all cases, the goal tolerance was set to $\epsilon_{\mathrm{goal}} = 0.2$~m.
For Cases 2 and 3, the characteristic length used for the manipulability normalization was set to $\ell_{\mathrm{c}} = 0.87$~m, corresponding to the maximum kinematic reach of a limb.
Using the normalized manipulability measures, the thresholds were set to $\epsilon_{\mathrm{m}} = 0.1$ for the limbs and $\epsilon_{\mathrm{b}} = 0.001$ for the base.

In addition to these comparative cases, an additional case was conducted to evaluate the planner under a sparser distribution of graspable points.
In this case (Case~4), the same planner settings as in Case~2 were used, while the environment was modified such that the total number of graspable points was reduced to 50\% of that in the original environment, including the removal of the floor graspable points located between the start and goal regions.

\subsection{Result and Discussion}

\fig{fig:case_study} illustrates the planned base paths and foothold sequences for Cases 1 through 3.
\tab{tab:case_metrics} summarizes the quantitative metrics of the generated paths, including the number of expanded nodes, total base translation, total steps to the goal, minimum limb and base manipulability measures, and computational time.
Here, $w_{\mathrm{m},\min}$ denotes the minimum normalized limb manipulability over all limbs along the planned path, whereas $w_{\mathrm{b},\min}$ denotes the minimum normalized base manipulability.

Comparing Case~1 and Case~2 highlights the impact of manipulability-based node selection.
As shown in \fig{fig:case_study} and \tab{tab:case_metrics}, introducing the manipulability constraints alters the planned path and foothold sequence.
In Case~1, $w_{\mathrm{m},\min}=0.07$ and $w_{\mathrm{b},\min}=0.0008$, both of which are below the respective thresholds of $\epsilon_{\mathrm{m}}=0.1$ and $\epsilon_{\mathrm{b}}=0.001$.
In contrast, Case~2 maintains both manipulability measures above their prescribed thresholds, demonstrating that the proposed node selection effectively excludes kinematically undesirable stances.
Case~2 required a larger number of expanded nodes and a longer computation time than Case~1.
Although the manipulability-based node selection locally prunes kinematically undesirable stances, it also restricts the feasible search space and may eliminate shorter or lower-cost routes.
Consequently, the A$^*$ search may need to explore alternative branches before reaching the goal.
The increase in the total step count from six in Case~1 to seven in Case~2 is also consistent with this interpretation.

The comparison between Cases~2 and 3 evaluates the influence of the cost function design.
As shown in \fig{fig:case_study}, changing the cost function resulted in different base paths and foothold sequences, demonstrating that the planner adapts its solution according to the specified optimization objective.
As summarized in \tab{tab:case_metrics}, Case~2 achieves a shorter total base translation than Case~3, whereas Case~3 reduces the number of locomotion steps.
The influence of the cost-function design is expected to become more pronounced in environments with different scales or more diverse distributions of graspable points.
While this study considers a relatively small-scale environment representing a space station module, evaluation under a wider range of environmental conditions remains an important topic for future work.

\begin{figure}[t]
    \centering
    \centerline{\includegraphics[width=0.8\linewidth]{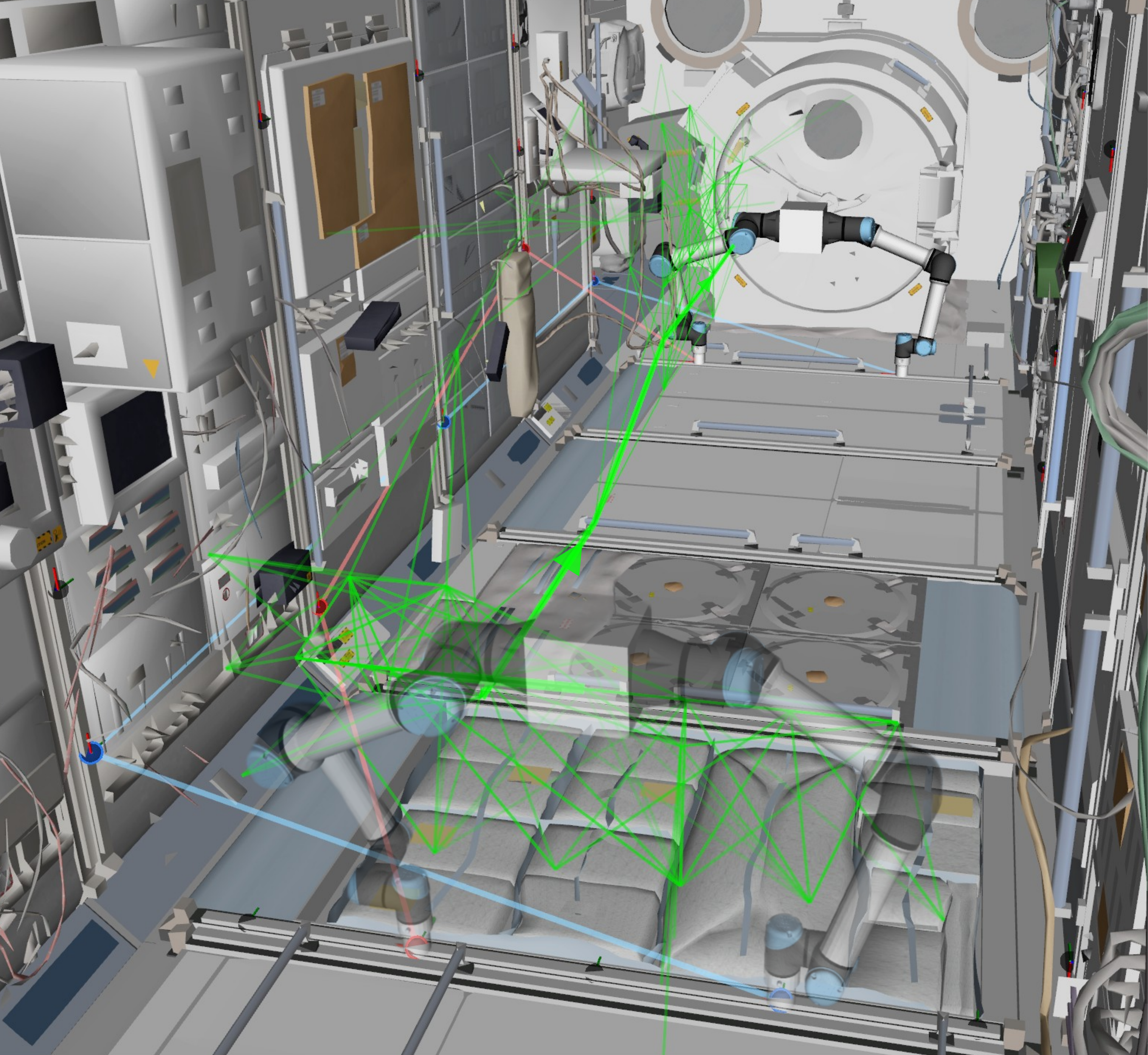}}
    \footnotesize
    - Manipulability-based node selection: yes\\
    - Cost: base translation\\
    \vspace{2mm}
    {\it Case 4}
    \caption{Cost-optimal paths and foothold sequences for a multi-limbed intra-vehicular robot on reduced graspable points of Case 4.}
    \label{fig:case4}
\end{figure}
Next, the result for {\it Case~4}, in which the number of graspable points was reduced, is shown in \fig{fig:case4}.
The planned path initially moves laterally before proceeding toward the goal, indicating that the planner selects an alternative route according to the sparser distribution of available graspable points.
The resulting plan requires 7 locomotion steps, with a total base translation of 5.0~m.
These results demonstrate that the proposed planner can generate a feasible path and foothold sequence even in an environment with a reduced number of graspable points.

\section{Dynamic Simulation} \label{sec:simulation}
\begin{figure*}[t]
    \centering
    \centerline{\includegraphics[width=\linewidth]{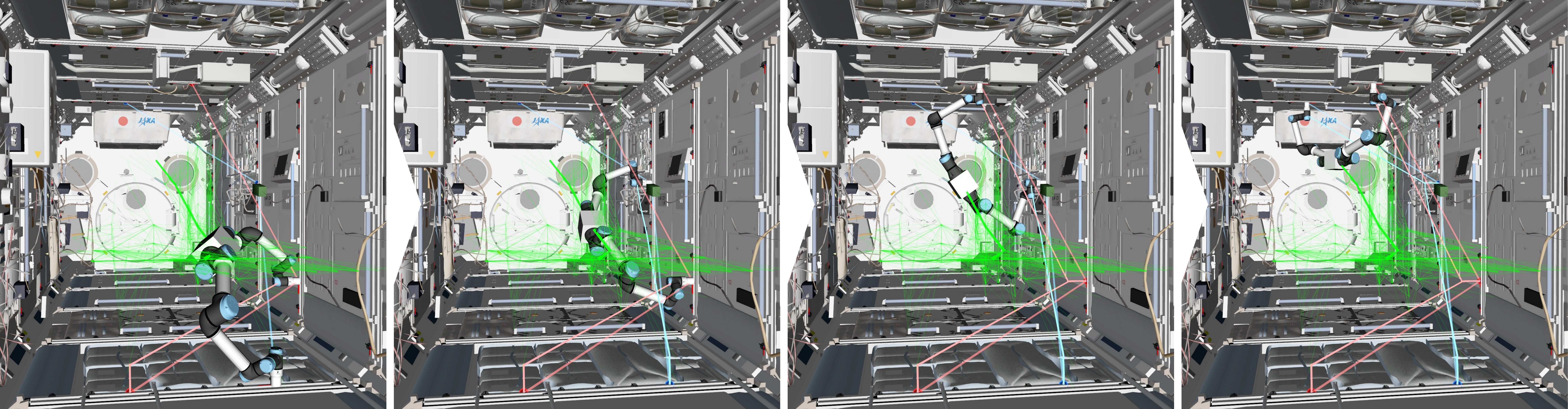}}
    \caption{Dynamic simulation of MLIVR locomotion in an ISS module using the proposed path and foothold planner. Straight lines indicate the planned base path (green) and foothold sequence (red and blue), whereas bright red and blue lines represent the actual trajectories tracked by the robot controller. The motion between successive planned stances is interpolated using quintic polynomial trajectories.
\label{final-sim}}
\end{figure*}
\begin{figure}[t]
    \centering
    \centerline{\includegraphics[width=\linewidth]{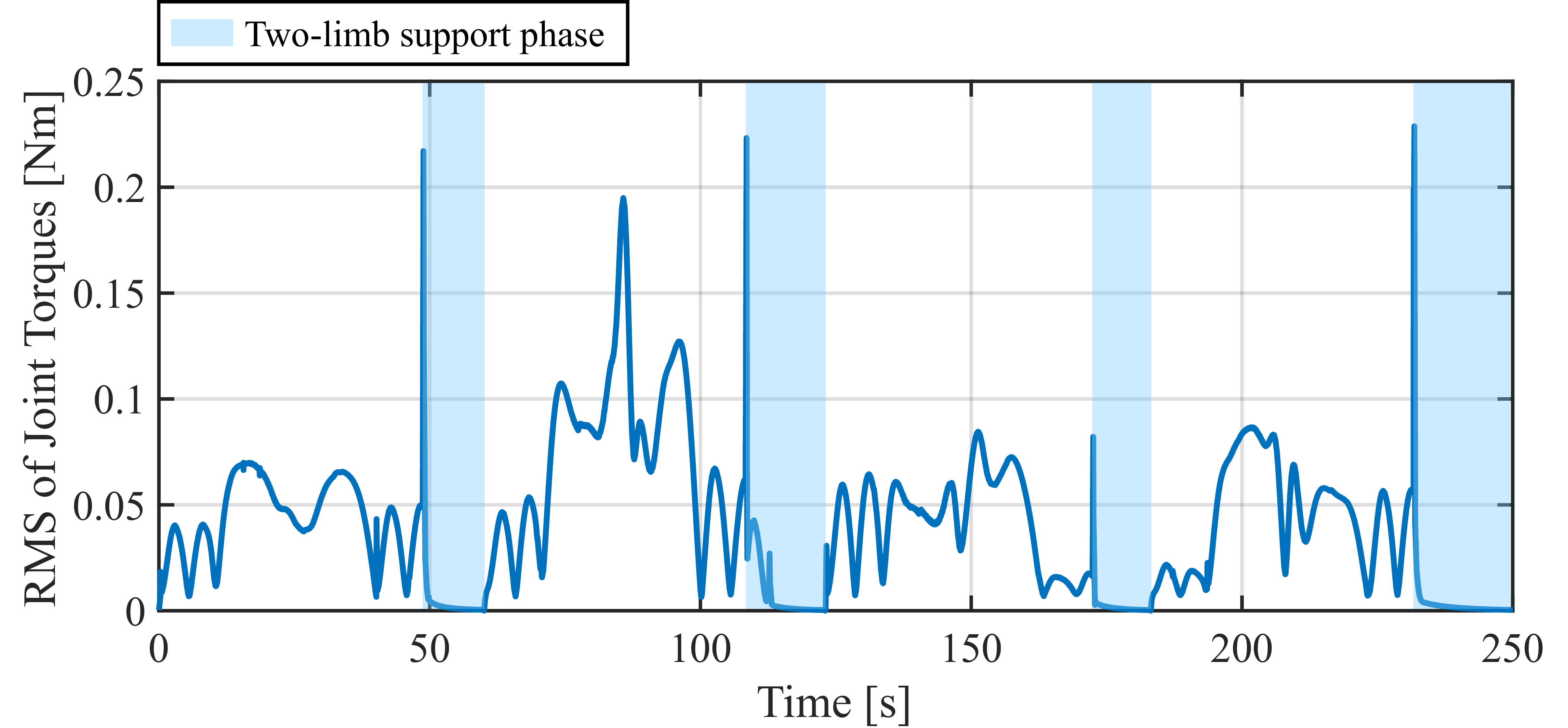}}
    \vspace{-2mm}
    \caption{Time history of the root mean square (RMS) of the whole-body joint torques.}
    \label{fig:torque-rms}
\end{figure}
Finally, to validate the dynamic feasibility of the generated path and foothold sequences, a dynamic simulation was performed using MuJoCo ver.~3.4, a high-performance physics engine.
For this validation, we adopted the locomotion plan generated using the step count cost with manipulability-based node selection.
All other planner settings were identical to those used in Section~\ref{sec:evaluation}.

\subsection{Controller and Physics Settings}

Once the proposed planner determines the sequence of base poses and footholds, continuous trajectories must be generated for the actual locomotion.
For each step, the trajectory of the swing limb's end-effector is generated using a quintic polynomial, with boundary conditions that set the velocity and acceleration to zero at both the start and end points.
To avoid impacts between the end-effector and the environment during disengagement and re-grasping, departure and approach phases are introduced before and after the swing phase, respectively.
During these phases, the end-effector moves away from or toward the grasping surface along its surface-normal direction.
Simultaneously, the base trajectory is interpolated using spherical linear interpolation for orientation and a quintic polynomial for translation with zero velocity and acceleration imposed at both endpoints.
In this study, to reduce the effects of reaction forces caused by dynamic motion, quasi-static transitions are assumed, and the duration of each locomotion step is set to 50.0~s.
Each step consists of a 10.0-s departure phase, a 30.0~s swing phase, and a 10.0~s approach phase.
The robot tracks the generated trajectories using a joint-space PD controller with proportional and derivative gains of
$k_{\mathrm{p}}=100$ and $k_{\mathrm{d}}=50$, respectively.

For the dynamics calculation, interactions between the robot and the environment are modeled using virtual spring-damper systems.
For collisions, repulsive forces are generated according to the penetration depth and relative velocity between the contacting geometries.
For grasping, restoring forces and torques are applied to maintain the end-effector at the target grasp pose, thereby representing a firm but compliant grasp.

\subsection{Result and Discussion}

As shown in \fig{final-sim}, the robot successfully accomplishes the complex transition from the floor to the ceiling without kinematic failure under the simulated microgravity conditions.
This result demonstrates that the path and foothold sequence generated by the proposed planner can be converted into continuous whole-body motion while maintaining the required environmental contacts.

To further evaluate the physical behavior during locomotion, \fig{fig:torque-rms} shows the time history of the root mean square (RMS) of the whole-body joint torques.
Distinct torque peaks are observed when the swing limb establishes contact with the environment at the end of each approach phase.
These peaks are attributed to the transient interaction forces generated when the end-effector engages with the grasp model.
In contrast, the joint torques remain relatively small during the slow departure and swing motions, which is consistent with the quasi-static transition assumed in this study.
Although the simulation confirms the feasibility of executing the planned stance transitions under the assumed contact model, the long step duration intentionally suppresses dynamic reaction forces.
Therefore, locomotion at more practical speeds will require trajectory generation and control that explicitly account for inertial reaction forces and contact-transition dynamics.

\section{Conclusion}\label{conclusion}

In this paper, a graph-based framework for the simultaneous planning of locomotion paths and footholds has been presented for the three-dimensional locomotion of a multi-limbed intra-vehicular robot (MLIVR) in space station environments.
The proposed planner explicitly considers the discrete distribution of graspable interfaces and generates cost-optimal locomotion plans while satisfying kinematic feasibility and manipulability constraints.
Comparative case studies demonstrated that manipulability-based node selection effectively excludes kinematically undesirable stances and that different cost functions produce distinct locomotion strategies consistent with their respective optimization objectives.
Furthermore, the planner successfully generated a feasible path and foothold sequence even when the total number of graspable points was reduced, demonstrating its applicability to sparser distributions of graspable interfaces.

Additionally, a 3D physics-based simulation in a realistic ISS-module environment was conducted to evaluate the feasibility of executing the generated locomotion plan.
The robot successfully completed the planned transition from the floor to the ceiling under simulated microgravity conditions, while the whole-body joint-torque response revealed transient loads associated with contact transitions.
These results demonstrate the applicability of the proposed graph-based planning framework and support its potential for autonomous intra-vehicular locomotion of MLIVRs in future orbital environments.

Despite these results, several extensions are required for operation in realistic intra-vehicular environments.
First, collision avoidance between the robot and its surroundings is not explicitly considered in the current planning framework.
In the confined and cluttered interior of a space station, collision constraints may significantly affect the feasibility of candidate stances and transitions and, consequently, the resulting locomotion route.
Therefore, collision checking should be incorporated into the graph-construction process to exclude infeasible nodes and transitions.
In addition, all graspable interfaces are currently assumed to be known a priori and continuously available.
In practice, however, these interfaces may be temporarily occupied or obstructed by onboard equipment or crew activities.
Extending the proposed framework with online replanning based on real time perception would enable the robot to adapt to such environmental changes and uncertainties.

Another important direction concerns the dynamic execution of the planned stance transitions.
Although the proposed method determines a global sequence of feasible stances, the transitions between successive stances were executed quasi-statically in this study to reduce the influence of dynamic effects, such as inertial reaction forces.
Achieving practical locomotion speeds will require these dynamic effects to be explicitly considered during motion generation and control.
Future work will therefore focus on dynamics-aware trajectory generation and control strategies that enable the robot to execute the planned stance sequence at realistic speeds while maintaining stable grasping and feasible whole-body motion.
Finally, experimental validation using real hardware remains an important direction for our future work.

\section*{Acknowledgment}
The authors would like to thank SpaceData Inc. for its invaluable support in developing a realistic world model of the ISS Japanese Experiment Module (JEM) for the simulation. The authors also thank Kazuki Takada for his contribution to the software development.

\bibliographystyle{IEEEtran}
\bibliography{reference.bib}

@article{ISS_Crew-time,
author = {Russell, James and others},
year = {2006},
month = {01},
pages = {130--136},
title = {Applying Analysis of International Space Station Crew-Time Utilization to Mission Design},
volume = {43},
journal = {Journal of Spacecraft and Rockets},
doi = {10.2514/1.16135}
}

@inproceedings{yamaguchi2025free,
  title={Free-flying crew cooperative robots on the ISS: A joint review of astrobee, CIMON, and int-ball operations},
  author={Yamaguchi, Seiko Piotr and others},
  booktitle={Proceedings of the 2025 International Conference on Space Robotics (iSpaRo)},
  pages={402--409},
  year={2025},
}

@article{hirano2024intball,
  title={Int-ball2: On-orbit demonstration of autonomous intravehicular flight and docking for image capturing and recharging},
  author={Hirano, Daichi and others},
  journal={IEEE Robotics \& Automation Magazine},
  volume={32},
  number={3},
  pages={76--87},
  year={2024},
  publisher={IEEE}
}

@article{smith2026astrobee,
  title={Astrobee: Free-flying robots for the international space station},
  author={Smith, Trey and others},
  journal={IEEE Transactions on Field Robotics},
  year={2026},
  publisher={IEEE}
}

@incollection{eisenberg2025cimon,
  author    = {Eisenberg, Till and others},
  title     = {CIMON -- The First Artificial Crew Assistant in Space},
  booktitle = {Aerospace Psychology and Human Factors: Applied Methods and Techniques},
  editor    = {Koglbauer, Ioana and Biede, Sonja},
  publisher = {Hogrefe Publishing},
  year      = {2025},
  pages     = {149--163},
  isbn      = {9781616766474}
}

@inproceedings{ssrms,
  title={{Flight 6A}: deployment and checkout of the space station remote manipulator system (SSRMS)},
  author={McGregor, Rod and Oshinowo, Layi},
  booktitle={Proceedings of the 6th International Symposium on Artificial Intelligence, Robotics and Automation in Space (i-SAIRAS)},
  year={2001}
}

@inproceedings{era,
  title={The european robotic arm: A high-performance mechanism finally on its way to space},
  author={Cruijssen, HJ and others},
  booktitle={The 42nd aerospace mechanism Symposium},
  year={2014}
}

@inproceedings{cssm,
  title={Overview of the Chinese space station manipulator},
  author={Li, Daming and Wang, Yaobing},
  booktitle={AIAA SPACE 2015 conference and exposition},
  pages={4540},
  year={2015}
}

@INPROCEEDINGS{R2_mobility,
  title={Robonaut 2 on the International Space Station: Status Update and Preparations for IVA Mobility},
  author={Thomas Ahlstrom and others},
  booktitle={Proceedings of the AIAA SPACE 2013 Conference and Exposition},
  year={2013},
  url={https://api.semanticscholar.org/CorpusID:110759145}
}

@inproceedings{yamaguchi2025towards,
  title={Towards the Automation in the Space Station: Feasibility Study and Ground Tests of a Multi-Limbed Intra-Vehicular Robot},
  author={Yamaguchi, Seiko Piotr and others},
  booktitle={2025 IEEE/SICE International Symposium on System Integration (SII)},
  pages={1095--1101},
  year={2025},
}

@article{bretl2006motion,
  title={Motion planning of multi-limbed robots subject to equilibrium constraints: The free-climbing robot problem},
  author={Bretl, Timothy},
  journal={The International Journal of Robotics Research},
  volume={25},
  number={4},
  pages={317--342},
  year={2006},
  publisher={SAGE Publications}
}

@inproceedings{albee2019motion,
  title={Motion planning for climbing mobility with implementation on a wall-climbing robot},
  author={Albee, Keenan and others},
  booktitle={2019 IEEE aerospace conference},
  pages={1--10},
  year={2019},
}

@inproceedings{uno2019gait,
  title={Gait planning for a free-climbing robot based on tumble stability},
  author={Uno, Kentaro and others},
  booktitle={2019 IEEE/SICE International Symposium on System Integration (SII)},
  pages={289--294},
  year={2019},
}

@article{xu2021contact,
  title={Contact sequence planning for hexapod robots in sparse foothold environment based on Monte-Carlo tree},
  author={Xu, Peng and others},
  journal={IEEE Robotics and Automation Letters},
  volume={7},
  number={2},
  pages={826--833},
  year={2021},
  publisher={IEEE}
}

@inproceedings{takada2023graph,
  title={Graph-based path/foothold planning and quantitative map evaluation for multi-limbed climbing robots},
  author={Takada, Kazuki and others},
  booktitle={2023 8th International Conference on Robotics and Automation Engineering (ICRAE)},
  pages={1--6},
  year={2023},
}

@inproceedings{rodriguez2024hybrid,
  title={Hybrid motion planner for a multi-armed robot performing on-orbit loco-manipulation tasks},
  author={Rodr{\'\i}guez, Ismael and others},
  booktitle={2024 IEEE Aerospace Conference},
  pages={1--9},
  year={2024},
}

@inproceedings{ohkami1999operational,
  title={Operational aspects of a super redundant space robot with reconfiguration and brachiating capability},
  author={Ohkami, Yoshiaki and others},
  booktitle={IEEE SMC'99 Conference Proceedings. 1999 IEEE International Conference on Systems, Man, and Cybernetics},
  volume={3},
  pages={178--183},
  year={1999},
}

@article{hayashi2000design,
  title={Design concept and system architecture of reconfigurable brachiating space robot},
  author={Hayashi, Ryoichi and others},
  journal={Journal of Robotics and Mechatronics},
  volume={12},
  number={4},
  pages={425--431},
  year={2000},
  publisher={Fuji Technology Press Ltd.}
}

@inproceedings{deremetz2021concept,
  title={Concept of operations and preliminary design of a modular multi-arm robot using standard interconnects for on-orbit large assembly},
  author={Deremetz, Mathieu and others},
  booktitle={72st International Astronautical Congress (IAC), Duba{\"\i}},
  year={2021}
}

@inproceedings{hoyt2013spiderfab,
  title={SpiderFab: An architecture for self-fabricating space systems},
  author={Hoyt, Robert P},
  booktitle={AIAA Space 2013 conference and exposition},
  pages={5509},
  year={2013}
}

@inproceedings{zhao2024model,
  title={Model Design and Concept of Operations of Standard Interface for On-orbit Construction},
  author={Zhao, Jingdong and others},
  booktitle={2024 IEEE International Conference on Robotics and Automation (ICRA)},
  pages={13487--13493},
  year={2024},
}

@article{yoshikawa1985manipulability,
  title={Manipulability of robotic mechanisms},
  author={Yoshikawa, Tsuneo},
  journal={The international journal of Robotics Research},
  volume={4},
  number={2},
  pages={3--9},
  year={1985},
  publisher={Sage Publications Sage CA: Thousand Oaks, CA}
}

@article{chiacchio1991global,
  title={Global task space manipulability ellipsoids for multiple-arm systems},
  author={Chiacchio, Pasquale and others},
  journal={IEEE Transactions on Robotics and Automation},
  volume={7},
  number={5},
  pages={678--685},
  year={1991}
}

@article{stocco1999use,
  title={On the use of scaling matrices for task-specific robot design},
  author={Stocco, Leo J and others},
  journal={IEEE Transactions on Robotics and Automation},
  volume={15},
  number={5},
  pages={958--965},
  year={1999},
  publisher={IEEE},
  doi = {10.1109/70.795800}
}

\end{document}